\documentclass[letterpaper]{article} 
\usepackage{aaai2027}  
\usepackage[hyphens]{url}  
\usepackage{graphicx} 
\usepackage{natbib}  
\usepackage{caption} 
\usepackage{algorithm}
\usepackage{algorithmic}
\usepackage{amsfonts}
\usepackage{amsmath}
\usepackage{amssymb}
\usepackage[table]{xcolor}
\usepackage{newfloat}
\usepackage{listings}
\DeclareCaptionStyle{ruled}{labelfont=normalfont,labelsep=colon,strut=off} 
\floatstyle{ruled}
\newfloat{listing}{tb}{lst}{}
\floatname{listing}{Listing}

\usepackage{booktabs}
\usepackage{multirow}
\title{ECAD: Expanding Class-Agnostic Detection Beyond Thing-Centric Objectness}
\author{
    Liang Wan\textsuperscript{\rm 1,2}, 
    Zixin Ren\textsuperscript{\rm 1},
    Yupeng Zhang\textsuperscript{\rm 1,2}\corresponding,
    Yuhan Wang\textsuperscript{\rm 1},
    Fangzhuo Gao\textsuperscript{\rm 1}
}
\affiliations{
	\textsuperscript{\rm 1}College of Intelligence and Computing, Tianjin University.\\
    \textsuperscript{\rm 2}Key Research Center for Surface Monitoring and Analysis of Relics, State Administration of Cultural Heritage.\\
}

\begin{document}

\maketitle

\begin{abstract}
Object detection is a fundamental task in visual perception, providing structured region representations for recognition, grounding, reasoning, and interaction. However, existing detection paradigms largely inherit a thing-centric notion of objectness, where detectors are mainly trained to localize discrete and countable object instances. Consequently, many semantically meaningful visual elements, such as sky, road, grassland, water, and sports courts, are often absorbed into the background despite their importance for scene understanding and spatial reasoning. In this paper, we formulate \textbf{Expanded Class-Agnostic Detection (ECAD)}, a new setting that aims to discover category-agnostic visual candidates beyond conventional thing-centric objects. To support this setting, we construct \textbf{BTCO-Bench}, a Beyond Thing-Centric Objectness benchmark with category-agnostic box annotations covering both real-world and cross-domain scenarios. We further propose \textbf{ECADet}, a lightweight DETR-based detector built upon a frozen DINOv3 encoder, and introduce \textbf{Geometry-Aware Expert Regression (GAER)} and \textbf{Prototype-Guided Query Modulation (PGQM)} to improve localization and objectness estimation for diverse visual elements, respectively. Extensive experiments show that ECADet consistently outperforms representative class-agnostic and proposal-based detectors on BTCO-Bench, demonstrating the effectiveness of expanded objectness discovery. \textit{Code and benchmark will be released.}
\end{abstract}


\section{Introduction}
Object detection is a fundamental component of visual perception, providing structured region representations for downstream recognition, reasoning, and interaction. Yet, what constitutes an object is often implicitly assumed. Mainstream datasets, such as COCO~\cite{lin2014microsoft}, LVIS~\cite{gupta2019lvis}, and PASCAL VOC~\cite{everingham2010pascal}, mainly annotate discrete, countable thing instances, while other semantically meaningful visual elements are treated as background. Although this thing-centric notion of objectness has driven substantial progress, it limits detector coverage in complex scenes.

\begin{figure}[t!]
	\centering
	\includegraphics[width=0.93\linewidth]{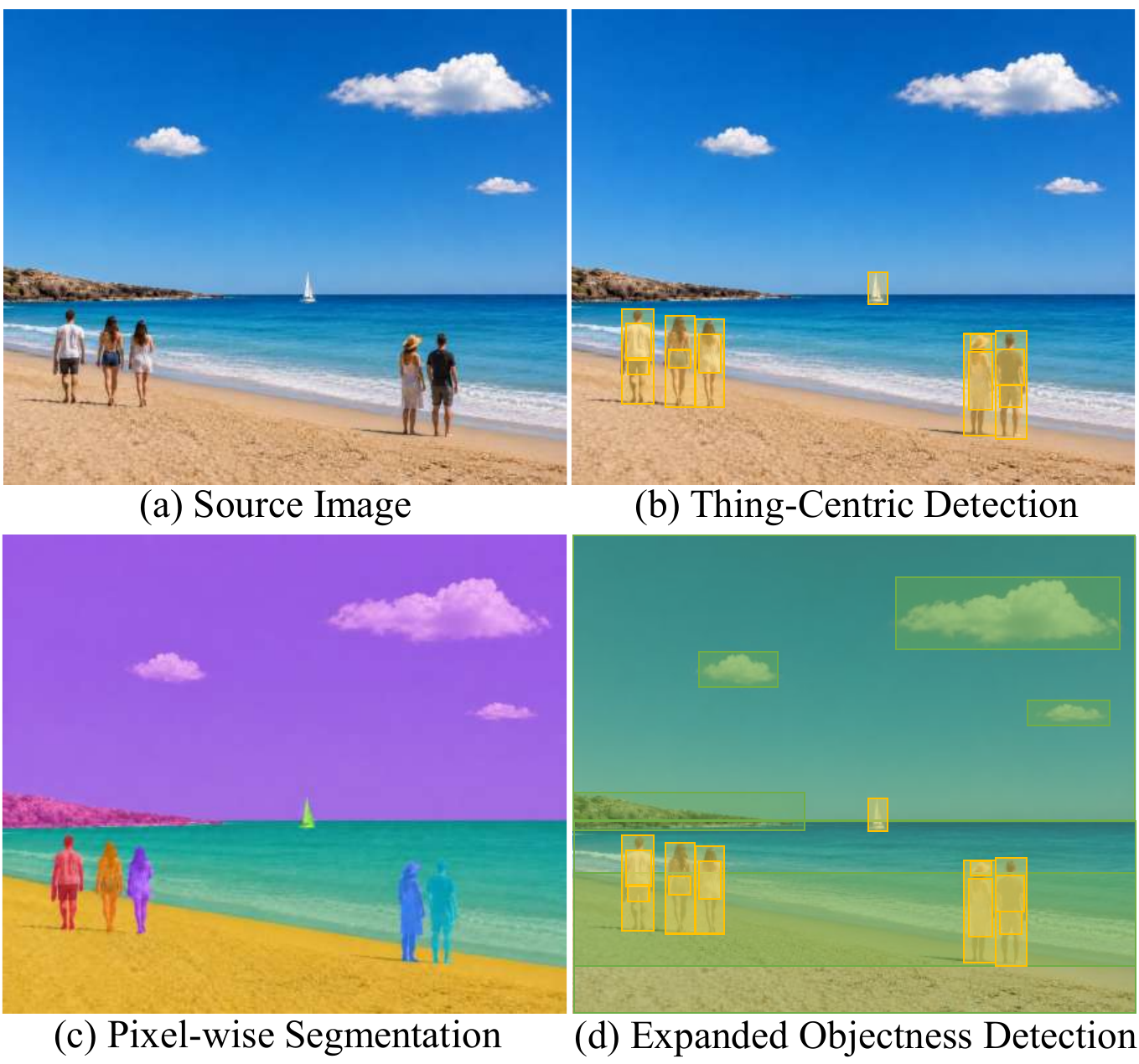}	
        \vspace{-7pt}
        \caption{Expanded Class-Agnostic Detection. Given a source image in (a), existing detectors in (b) are constrained by thing-centric objectness, while segmentation in (c) provides dense pixel-wise labels rather than sparse detection candidates. In contrast, ECAD in (d) expands the object discovery space by discovering both thing objects and semantically meaningful visual elements as category-agnostic boxes.}  
	\label{fig:motivation}
	\vspace{-18pt}
\end{figure}

As shown in Fig.~\ref{fig:motivation}, real-world scenes contain many visual elements that fall outside the conventional definition of object instances yet carry clear semantic, functional, or interactive value. Roads guide navigation, walls and floors define spatial layouts, while sky, water, grassland, and courts provide important scene context. Although often non-countable and less clearly bounded than common objects, these elements are not mere background; they serve as semantic anchors for scene understanding, spatial reasoning, embodied navigation, and multimodal grounding. Expressions such as people on the `grassland', buildings under the `sky', players at the center of the `court', and boats on the `sea' all rely on such anchors. If the visual front end detects only conventional thing objects, these references cannot be effectively grounded or used for reasoning. Thus, the limitation of existing detectors lies not only in which categories they can recognize, but also in which visual content they can discover as objects.

Recent advances in class-agnostic, open-vocabulary, and open-world detection have relaxed category constraints from different perspectives. Class-agnostic detection (CAD)~\cite{jaiswal2021class,maaz2022class,syuen2024dipex,tang2026prompt} learns general objectness without category-specific classification, open-vocabulary detection (OVD)~\cite{gu2021open,fu2025llmdet,zhang2026noovd} recognizes categories unseen during training, and open-world detection (OWD)~\cite{joseph2021towards,ma2025skdf} addresses unknown discovery and incremental learning. However, these paradigms still inherit thing-centric supervision from existing datasets. They expand the category space or improve category generalization, but rarely reconsider the fundamental question of what visual content should be discovered as objects. Consequently, many semantically or functionally meaningful visual elements beyond conventional object instances remain outside the object discovery space.

Motivated by these observations, we revisit the scope of objectness in CAD. Rather than asking which category a visual element belongs to, we ask \textbf{which visual content should be discovered as category-agnostic object candidates}. Conventional class-agnostic detectors mainly target thing instances, whereas our goal is to cover both common objects, such as humans, vehicles, and animals, and semantically meaningful elements often treated as background, including sky, sea, grassland, roads, and courts. \textit{Although semantic and panoptic segmentation also model such stuff-like regions, they perform dense pixel-wise labeling rather than sparse candidate discovery. Their flat output spaces are also less suited to representing overlapping or hierarchical candidates, such as a person and its parts.}
Importantly, our goal is neither to enlarge the detection vocabulary nor to convert segmentation labels into detection categories. The model still predicts category-agnostic candidate boxes without assigning class names. We therefore introduce \textbf{Expanded Class-Agnostic Detection (ECAD)}, which broadens objectness from thing-centric instances to a wider range of semantically meaningful visual elements while preserving the category-agnostic box prediction formulation.

A dedicated benchmark is needed to evaluate this setting. Existing detection benchmarks largely inherit thing-centric annotations, leaving many semantically meaningful non-thing elements unlabeled and treated as background. They therefore cannot reliably assess a detector’s ability to discover visual content beyond conventional object instances. To address this gap, we construct \textbf{BTCO-Bench}, a category-agnostic benchmark for Beyond Thing-Centric Objectness, covering both real-world and cross-domain scenarios. Further details are provided in the \textbf{BTCO-Bench} section.

We further propose \textbf{ECADet}, an Expanded Class-Agnostic Detector for discovering semantically meaningful visual elements beyond thing-centric objectness. Built on RF-DETR, ECADet freezes the large-scale pretrained DINOv3 encoder to preserve its transferable visual representations and reduce overfitting to limited training data, while training only a lightweight projector, Transformer decoder, and two objectness-oriented modules. \textbf{Geometry-Aware Expert Regression (GAER)} improves localization for elements with diverse shapes and boundary ambiguity, while \textbf{Prototype-Guided Query Modulation (PGQM)} injects structural foreground priors into object queries. This lightweight design broadens the object discovery space without retraining the visual foundation model, providing richer category-agnostic proposals for downstream recognition, grounding, segmentation, and multimodal reasoning.

Our main contributions are summarized as follows:
\begin{itemize}
\item 
We revisit the implicit assumption of thing-centric objectness in existing detection paradigms and introduce \textbf{ECAD}, which extends category-agnostic object discovery from conventional thing instances to broader semantically meaningful visual elements.
\item 
We construct \textbf{BTCO-Bench}, a Beyond Thing-Centric Objectness benchmark with category-agnostic bounding-box annotations across real-world and cross-domain scenarios. We further propose \textbf{ECADet}, a lightweight DETR-based baseline equipped with \textbf{Geometry-Aware Expert Regression (GAER)} and \textbf{Prototype-Guided Query Modulation (PGQM)}.
\item 
Extensive experiments show that ECADet consistently outperforms representative class-agnostic and proposal-based detectors on BTCO-Bench, demonstrating its effectiveness in discovering visual content beyond thing-centric objectness.
\end{itemize}

\section{Related Work}

\textbf{Class-agnostic detection (CAD)} localizes object candidates without assigning category labels, improving generalization to unseen categories and domains. Early work decoupled localization from category prediction~\cite{jaiswal2021class}, while later methods leveraged vision-language supervision and large-scale grounding data to improve proposal quality. MAVL~\cite{maaz2022class} uses box-level supervision from vision-language datasets, DiPEx~\cite{syuen2024dipex} expands textual queries to better cover the object concept space, and PF-RPN~\cite{tang2026prompt} progressively refines a learnable visual embedding for prompt-free proposal generation across unseen domains. However, these methods still learn objectness from thing-centric annotations of discrete, countable, and nameable entities, leaving their discovery space constrained by existing detection datasets. In contrast, we extend category-agnostic discovery from conventional thing instances to semantically meaningful non-thing elements.

\textbf{Open-vocabulary detection (OVD) and open-world detection (OWD)} relax the category constraints of closed-set detection from complementary perspectives. OVD methods, such as ViLD~\cite{gu2021open}, LLMDet~\cite{fu2025llmdet}, and NoOVD~\cite{zhang2026noovd}, associate visual regions with open-ended textual concepts to recognize categories unseen during training. OWD methods, such as OWD~\cite{joseph2021towards} and SKDF~\cite{ma2025skdf}, further address unknown object discovery and incremental category learning during deployment. However, both paradigms largely inherit the object definition of standard detection datasets, where detection targets remain thing-centric instances. Consequently, semantically or functionally meaningful elements outside conventional instance annotations are still treated as background. In contrast, our goal is not to recognize more category names, but to broaden the visual content included in category-agnostic object discovery.

\textbf{Semantic, panoptic, and open-vocabulary segmentation} provide dense scene parsing through pixel-level prediction. Representative works include FCN~\cite{long2015fully} for semantic segmentation, the panoptic segmentation formulation~\cite{kirillov2019panoptic}, and LSeg~\cite{li2022language} and OpenSeg~\cite{ghiasi2022scaling} for open-vocabulary segmentation. Benchmarks such as ADE20K~\cite{zhou2017scene} and COCO-Stuff~\cite{caesar2018coco} naturally cover stuff categories including sky, road, ground, and wall. However, these tasks differ fundamentally from our setting. Segmentation produces dense pixel-level outputs, typically assigning each pixel a predefined or text-specified semantic label, whereas our task discovers a sparse set of category-agnostic candidate boxes with objectness scores. Thus, our work is complementary to segmentation: rather than determining the semantic label of each pixel, we identify which semantically meaningful visual elements should be explicitly proposed as object candidates.

\section{BTCO-Bench: Beyond Thing-Centric Objectness Benchmark}

\subsection{Motivation}
Existing detection benchmarks are not designed for ECAD. Datasets such as COCO and LVIS largely annotate countable thing instances, leaving many semantically meaningful, especially stuff-like, elements unlabeled and treated as background. Consequently, they cannot assess a detector's ability to discover visual content beyond conventional object instances.
Moreover, this expanded notion of objectness should generalize from standard natural images to artistic and degraded domains. Yet existing cross-domain benchmarks likewise lack category-agnostic box annotations for such elements. To address this gap, we construct \textbf{BTCO-Bench}, a Beyond Thing-Centric Objectness benchmark for evaluating category-agnostic discovery of both conventional thing instances and semantically meaningful visual elements beyond the thing-centric annotation space across real-world and cross-domain scenarios.

\subsection{Benchmark Construction}

BTCO-Bench consists of two complementary splits: BTCO-Real and BTCO-XDomain.
\textbf{BTCO-Real} is built upon LVIS, which contains diverse real-world scenes and everyday objects. We augment the LVIS minival split with bounding-box annotations for semantically meaningful elements beyond its original annotation space. These include stuff-like regions, such as sky, roads, grassland, water, mountains, and courts, as well as previously unannotated object-like elements, such as police lights, rocks, flowers, trees, windows, billboards, and signs. We further sample 1,000 LVIS training images and annotate them under the same protocol, forming a compact training split for expanded objectness learning. All regions are represented as category-agnostic boxes with a unified foreground label, \textit{i.e.}, \textit{object}. BTCO-Real thus evaluates the discovery of conventional thing instances together with previously unannotated object-like and semantically meaningful stuff-like elements.
\textbf{BTCO-XDomain} evaluates the cross-domain generalization of expanded objectness under domain shifts. It comprises three subsets: artistic images from MSOSB~\cite{zhang2024rethinking}, curated by removing style-transferred and implausible AI-generated samples and retaining watercolor, cartoon, and oil-painting styles; monochrome and colored pencil drawings collected from the Internet; and adverse-condition images from SDGOD~\cite{wu2022single}, excluding style-transferred samples and covering \texttt{Daytime foggy}, \texttt{Dusk rainy}, and \texttt{Night clear}. Compared with BTCO-Real, this split covers \textbf{seven} conditions involving artistic abstraction, sketch-like contours, adverse weather, and illumination changes, enabling evaluation of expanded objectness in unseen domains.

Five trained annotators participated in image collection, data cleaning, bounding-box annotation, and annotation standardization. All annotations follow the COCO format and assign each region the unified foreground label \textit{object}. Category names are not used for training or evaluation, ensuring strict adherence to the class-agnostic detection setting. The statistics of BTCO-Bench are summarized in Table~\ref{tab:btco_statistics}. The table reports the number of images and annotated boxes in each split, together with their scale distributions, highlighting the diverse object sizes covered by the benchmark.

\subsection{Design Principles and Characteristics}
BTCO-Bench follows four key design principles. 1) It broadens objectness coverage from conventional thing instances to semantically meaningful non-thing elements. 2) All annotations are category-agnostic, focusing evaluation on objectness estimation and localization rather than category recognition. 3) The benchmark covers elements with diverse scales, aspect ratios, shapes, and boundary ambiguity; non-thing elements are often less compact and less clearly bounded than regular objects, posing greater localization challenges. 4) BTCO-XDomain introduces substantial appearance shifts through artistic images, pencil drawings, and adverse-condition scenes, enabling evaluation of objectness generalization beyond standard natural-image domains.
More details are provided in the \textit{\textbf{supplementary material}}.

\begin{table}[t!]
\centering
\scriptsize
\setlength{\tabcolsep}{1.8pt}
\caption{Statistics of BTCO-Bench.}
\label{tab:btco_statistics}
\vspace{-10pt} 
\resizebox{\linewidth}{!}{
\begin{tabular}{lccccccc}
\toprule
\multirow{2}{*}{Split} 
& \multirow{2}{*}{Source} 
& \multirow{2}{*}{Domain} 
& \multirow{2}{*}{Images} 
& \multirow{2}{*}{Boxes} 
& \multicolumn{3}{c}{Box Size} \\
\cmidrule(lr){6-8}
& & & & & Small & Medium & Large \\
\midrule
BTCO-Real (Train) & LVIS train & Real & 1,000 & 28,242 & 8,827 & 10,832 & 8,583 \\
BTCO-Real (Test) & LVIS minival & Real & 5,000 & 145,631 & 48,939 & 53,296 & 43,396 \\
BTCO-XDomain & MSOSB/Web/SDGOD & Cross-domain & 18,639 & 134,143 & 10,105 & 38,144 & 85,894 \\
\bottomrule
\end{tabular}}
\vspace{-17pt} 
\end{table}

\begin{figure*}[t!]
	\centering
	\includegraphics[width=1.0\linewidth]{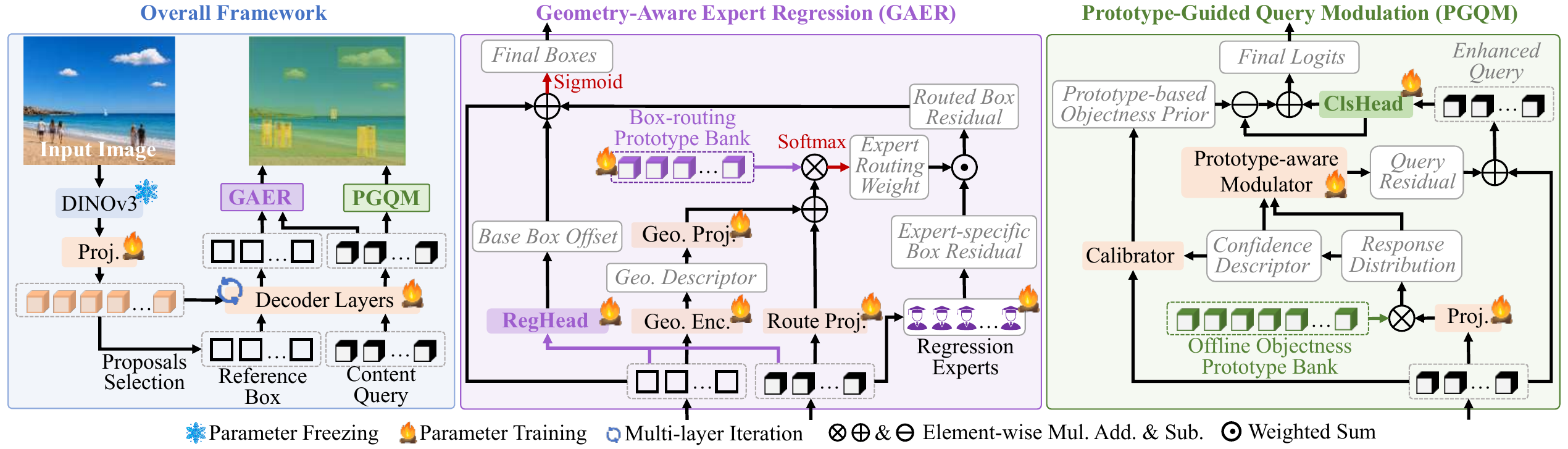}
	\vspace{-17pt}
    \caption{Overview of the proposed ECADet framework. A frozen DINOv3 extracts dense visual features, followed by a lightweight DETR decoder. Geometry-Aware Expert Regression (GAER) routes queries to specialized regression experts for diverse geometric patterns, while Prototype-Guided Query Modulation (PGQM) enhances query representations with offline objectness prototypes for more reliable objectness prediction.} 
	\label{fig:overall}
	\vspace{-15pt}
\end{figure*}

\section{The Proposed Method}

\subsection{Problem Formulation}

We consider the \textbf{Expanded Class-Agnostic Detection (ECAD)} problem. Given an input image $x$, the ground-truth set is
$\mathcal{Y}=\{\mathbf{b}_i\}_{i=1}^{N}$,
where all annotated regions share a unified foreground label regardless of their semantic categories. The model predicts a set of category-agnostic candidates:
\begin{equation}
\footnotesize
\hat{\mathcal{Y}}
=
\{(\hat{s}_j,\hat{\mathbf{b}}_j)\}_{j=1}^{Q},
\end{equation}
where $\hat{s}_j$ denotes the objectness score and
$\hat{\mathbf{b}}_j \in [0,1]^4$ is the normalized bounding box of the $j$-th candidate. Unlike closed-set and open-vocabulary detection, ECAD does not assign category labels to predicted regions. Instead, it treats both conventional thing instances and semantically meaningful non-thing elements as foreground candidates. The goal is to learn an expanded objectness function that discovers both conventional thing instances and semantically meaningful visual elements beyond the thing-centric annotation space.

\subsection{Overview of the Method}
Our framework builds on RF-DETR, as shown in Fig.~\ref{fig:overall}. A frozen DINOv3 encoder extracts visual features, followed by a lightweight projector and Transformer decoder that produce query features $\mathbf{Q}^{l}=\{\mathbf{q}^{l}_{i}\}_{i=1}^{N}$ and reference boxes $\mathbf{R}^{l}=\{\mathbf{r}^{l}_{i}\}_{i=1}^{N}$ at each decoder layer. Although all foreground regions share a unified label in ECAD, they remain highly diverse in appearance and geometry. We therefore introduce two complementary modules. Geometry-Aware Expert Regression (GAER) routes queries to specialized regression experts based on their features and reference-box geometry, while Prototype-Guided Query Modulation (PGQM) modulates query representations with offline DINOv3-derived objectness prototypes to improve objectness estimation. \textit{Both modules retain the original RF-DETR heads as the primary prediction paths and introduce lightweight residual refinements for ECAD.}

\subsection{Geometry-Aware Expert Regression (GAER)}

In ECAD, all foreground regions share a unified label despite substantial geometric diversity in scale, aspect ratio, shape, and localization difficulty. A shared regression head must model these heterogeneous patterns with a single mapping, limiting region-specific box refinement. We therefore introduce Geometry-Aware Expert Regression (GAER), which routes each query to specialized regression experts based on its representation and reference-box geometry.

Specifically, given the query features
$\mathbf{Q}^{l}=\{\mathbf{q}^{l}_{i}\}_{i=1}^{N}$
and the corresponding reference boxes
$\mathbf{R}^{l}=\{\mathbf{r}^{l}_{i}\}_{i=1}^{N}$
at the $l$-th decoder layer, we retain the original regression pathway of RF-DETR.
For the $i$-th query, the \textbf{original regression head (RegHead)} predicts a \textit{\textbf{Base Box Offset}}:
\begin{equation}
\footnotesize
\Delta \mathbf{b}^{l}_{\mathrm{base},i}
=
\mathrm{RegHead}(\mathbf{q}^{l}_{i}),
\label{eq:base_box_offset}
\end{equation}
where $\mathbf{q}^{l}_{i}\in\mathbb{R}^{d}$ denotes the $i$-th query feature,
$d$ is the hidden dimension, and
$\Delta \mathbf{b}^{l}_{\mathrm{base},i}\in\mathbb{R}^{4}$ is the base box refinement predicted by the original regression head.

Meanwhile, we encode the reference-box geometry for routing. Let
$\tilde{\mathbf{r}}^{l}_{i}=(c_x,c_y,w,h)$ denote the normalized reference box obtained from the decoder reference state. We construct a 6-dimensional \textit{\textbf{Geometry Descriptor}}:
\begin{equation}
\footnotesize
\mathbf{g}^{l}_{i}
=
\psi(\tilde{\mathbf{r}}^{l}_{i})
=
[c_x, c_y, \log w, \log h, \log(wh), \log(w/h)],
\label{eq:box_geometry_encoder}
\end{equation}
where $\psi(\cdot)$ denotes the box \textbf{Geometry Encoder}, capturing center, scale, area, and aspect ratio.

To determine which \textbf{Regression Experts} are suitable for the current query, we jointly use the query feature and its reference-box geometry. The query feature and \textit{\textbf{Geometry Descriptor}} are projected into the same routing space and then fused by addition:
\begin{equation}
\footnotesize
\mathbf{t}^{l}_{i}
=
f_{\mathrm{route}}(\mathbf{q}^{l}_{i})
+
f_{\mathrm{geo}}(\mathbf{g}^{l}_{i}),
\label{eq:routing_feature}
\end{equation}
where $\mathbf{t}^{l}_{i}\in\mathbb{R}^{d}$ is the routing feature.
Here, $f_{\mathrm{route}}(\cdot)$ is \textbf{Route Projecter}, a two-layer MLP that maps the query feature to the routing space, and
$f_{\mathrm{geo}}(\cdot)$ is \textbf{Geometry Projecter}, a two-layer MLP that maps the 6-dimensional \textit{\textbf{Geometry Descriptor}} to the same space.

We maintain a learnable \textbf{Box-routing Prototype Bank}
$\mathbf{P}_{\mathrm{box}}=\{\mathbf{p}_{k}\}_{k=1}^{K}$,
where $K$ is the number of \textbf{Regression Experts} and each prototype $\mathbf{p}_{k}\in\mathbb{R}^{d}$ represents a latent localization pattern.
The routing score between the $i$-th query and the $k$-th prototype is computed by cosine similarity:
\begin{equation}
\footnotesize
a^{l}_{i,k}
=
\mathrm{sim}(\mathbf{t}^{l}_{i}, \mathbf{p}_{k}),
\label{eq:routing_score}
\end{equation}
where $\mathrm{sim}(\cdot,\cdot)$ denotes cosine similarity.
The \textit{\textbf{Expert Routing Weight}} is obtained by a temperature-scaled softmax:
\begin{equation}
\footnotesize
w^{l}_{i,k}
=
\frac{
\exp(a^{l}_{i,k}/\tau_{\mathrm{box}})
}{
\sum_{j=1}^{K}\exp(a^{l}_{i,j}/\tau_{\mathrm{box}})
},
\label{eq:routing_weight}
\end{equation}
where $w^{l}_{i,k}$ is the weight assigned to the $k$-th \textbf{Regression Expert}, and
$\tau_{\mathrm{box}}$ is a temperature parameter.

Each \textbf{Regression Expert} predicts an \textit{\textbf{Expert-specific Box Residual}} from the query feature:
\begin{equation}
\footnotesize
\Delta \mathbf{b}^{l}_{\mathrm{expert},i,k}
=
E_{k}(\mathbf{q}^{l}_{i}),
\label{eq:expert_residual}
\end{equation}
where $E_{k}(\cdot)$ denotes the $k$-th \textbf{Regression Expert}, comprising a lightweight adapter and linear box regressor.

The final \textit{\textbf{Routed Box Residual}} is obtained by weighted aggregation over all experts:
\begin{equation}
\footnotesize
\Delta \mathbf{b}^{l}_{\mathrm{route},i}
=
\eta
\sum_{k=1}^{K}
w^{l}_{i,k}
\Delta \mathbf{b}^{l}_{\mathrm{expert},i,k},
\label{eq:routed_residual}
\end{equation}
where $\Delta \mathbf{b}^{l}_{\mathrm{route},i}$ denotes the \textit{\textbf{Routed Box Residual}}, with scale controlled by $\eta$.

Finally, the \textit{\textbf{Routed Box Residual}} is added to the \textit{\textbf{base regression result}} rather than replacing it. The final box prediction is obtained by refining the reference box with both the \textit{\textbf{base box offset}} and the \textit{\textbf{Routed Box Residual}}:
\begin{equation}
\footnotesize
\mathbf{b}^{l}_{i}
=
\sigma
\left(
\mathbf{r}^{l}_{i}
+
\Delta \mathbf{b}^{l}_{\mathrm{base},i}
+
\Delta \mathbf{b}^{l}_{\mathrm{route},i}
\right),
\label{eq:final_box}
\end{equation}
where $\sigma(\cdot)$ is the sigmoid and $\mathbf{b}^{l}_{i}$ is the final predicted box.

Overall, GAER preserves the original regression head's stable localization while augmenting it with query-adaptive expert residuals, enabling shared localization knowledge and selective modeling of heterogeneous geometric patterns across class-agnostic foreground regions.

\subsection{Prototype-Guided Query Modulation (PGQM)}

In parallel with the regression branch's modeling of geometric diversity, the classification branch handles the appearance and contextual diversity of class-agnostic foreground regions. Since ECAD reduces classification to objectness estimation without category supervision, a single classifier may struggle to capture diverse foreground modes. We therefore propose Prototype-Guided Query Modulation (PGQM), which uses objectness prototypes as structural foreground priors to modulate query features before classification.

We first construct an \textbf{Offline Objectness Prototype Bank} from the training set. A frozen DINOv3 encoder extracts dense features, from which foreground regions are cropped using ground-truth boxes and pooled into object-level embeddings for clustering into objectness prototypes. Derived from self-supervised representations and foreground annotations, these prototypes provide diverse category-agnostic priors for foreground appearance.

Given the query features
$\mathbf{Q}^{l}=\{\mathbf{q}^{l}_{i}\}_{i=1}^{N}$,
we project each query feature into the prototype matching space:
\begin{equation}
\footnotesize
\mathbf{z}^{l}_{i}=f_{\mathrm{proj}}(\mathbf{q}^{l}_{i}),
\label{eq:cls_projector}
\end{equation}
where $f_{\mathrm{proj}}(\cdot)$ is a lightweight \textbf{Projector}. We then compute the cosine similarity between $\mathbf{z}^{l}_{i}$ and the offline objectness prototype bank
$\mathbf{P}_{\mathrm{obj}}=\{\mathbf{p}_{m}\}_{m=1}^{M}$:
\begin{equation}
s^{l}_{i,m}=\mathrm{sim}(\mathbf{z}^{l}_{i},\mathbf{p}_{m}).
\label{eq:cls_similarity}
\end{equation}

The prototype \textit{\textbf{Response Distribution}} is obtained by temperature-scaled softmax:
\begin{equation}
\footnotesize
\pi^{l}_{i,m}
=
\frac{\exp(s^{l}_{i,m}/\tau_{\mathrm{obj}})}
{\sum_{n=1}^{M}\exp(s^{l}_{i,n}/\tau_{\mathrm{obj}})},
\label{eq:cls_proto_prob}
\end{equation}
where $\tau_{\mathrm{obj}}$ is a temperature parameter.

Instead of using prototypes as an independent classifier, we derive a
\textit{\textbf{Confidence Descriptor}} from the similarity scores and the \textit{\textbf{Response Distribution}}:
\begin{equation}
\footnotesize
\mathbf{c}^{l}_{i}
=
\Phi(\mathbf{s}^{l}_{i}, \boldsymbol{\pi}^{l}_{i}, \mathbf{q}^{l}_{i}),
\label{eq:cls_confidence_descriptor}
\end{equation}
where $\mathbf{s}^{l}_{i}$ denotes the raw prototype similarity scores and
$\boldsymbol{\pi}^{l}_{i}$ denotes the normalized prototype response distribution.
The \textit{\textbf{Confidence Descriptor}} $\mathbf{c}^{l}_{i}$ contains confidence-related statistics, including normalized entropy, top-1/top-2 response margin, pooled prototype response, maximum prototype similarity, query feature norm, and the derived confidence score. These statistics indicate whether a query matches foreground prototypes confidently and compactly, or responds ambiguously to multiple prototypes.

The \textbf{Prototype-aware Modulator} uses both the \textit{\textbf{Response Distribution}} and the \textit{\textbf{Confidence Descriptor}} to generate a \textit{\textbf{Query Residual}}. Specifically, we first construct an entropy-aware \textit{\textbf{Response Distribution}}:
\begin{equation}
\footnotesize
\bar{\boldsymbol{\pi}}^{l}_{i}
=
(1-H^{l}_{i})\boldsymbol{\pi}^{l}_{i}
+
H^{l}_{i}\mathbf{u},
\label{eq:entropy_aware_response}
\end{equation}
where $H^{l}_{i}$ is the normalized entropy of $\boldsymbol{\pi}^{l}_{i}$ and $\mathbf{u}$ the uniform prototype distribution. Under uncertain responses, this operation downweights unreliable assignments.

The modulator then maps $\bar{\boldsymbol{\pi}}^{l}_{i}$ with a two-layer MLP and maps the normalized query direction with a linear layer. Their sum is normalized and filtered by a confidence-aware gate:
\begin{equation}
\footnotesize
\Delta \mathbf{q}^{l}_{i}
=
\mathcal{M}(\mathbf{q}^{l}_{i}, \bar{\boldsymbol{\pi}}^{l}_{i}, \mathbf{c}^{l}_{i}),
\label{eq:cls_query_residual}
\end{equation}
where $\mathcal{M}(\cdot)$ denotes the \textbf{Prototype-aware Modulator}. A separate two-layer MLP predicts the gate using the entropy-aware response distribution and confidence cues in $\mathbf{c}^{l}_{i}$, including margin, confidence score, entropy, and query norm.

The \textit{\textbf{Enhanced Query}} is obtained by residual modulation:
\begin{equation}
\footnotesize
\hat{\mathbf{q}}^{l}_{i}
=
\mathbf{q}^{l}_{i}
+
\lambda
\Delta \mathbf{q}^{l}_{i},
\label{eq:cls_enhanced_query}
\end{equation}
where $\lambda$ is a learnable modulation scale.  

The \textit{\textbf{Enhanced Query}} is fed into the \textbf{original classification head (ClsHead)}:
\begin{equation}
\footnotesize
o^{l}_{\mathrm{base},i}
=
\mathrm{ClsHead}(\hat{\mathbf{q}}^{l}_{i}).
\label{eq:cls_base_logit}
\end{equation}

For conservative objectness calibration, we estimate a \textit{\textbf{Prototype-based Objectness Prior}}:
\begin{equation}
\footnotesize
o^{l}_{\mathrm{prior},i}
=
G_{\mathrm{obj}}(\mathbf{q}^{l}_{i},\boldsymbol{\phi}^{l}_{i}),
\label{eq:cls_objectness_prior}
\end{equation}
where $G_{\mathrm{obj}}(\cdot)$ is a low-capacity \textbf{Calibrator}. It computes a weighted combination of query norm, maximum prototype similarity, pooled prototype similarity, and prototype entropy, with positive learnable weights and a learnable bias.

The final objectness logit is obtained by bounded residual calibration:
\begin{equation}
\footnotesize
o^{l}_{i}
=
o^{l}_{\mathrm{base},i}
+
\beta \gamma^{l}_{i}
\tanh
\left(
o^{l}_{\mathrm{prior},i}
-
o^{l}_{\mathrm{base},i}
\right),
\label{eq:cls_final_logit}
\end{equation}
where $\gamma^{l}_{i}$ is a dynamic confidence gate predicted by a two-layer MLP from objectness prior, query norm, confidence, margin, and pooled similarity. $\beta$ is a learnable calibration scale,  
and $\tanh(\cdot)$ bounds the correction magnitude.

Overall, PGQM retains the original classification head as the main prediction path, using offline DINOv3-driven objectness prototypes only for confidence-aware query modulation and conservative logit calibration to improve objectness estimation.

\subsection{Training Objective and Implementation Details}

\textbf{Training objective.}
ECADet follows RF-DETR's set prediction objective. After Hungarian matching, it is optimized with classification, $\ell_1$ regression, and GIoU losses. GAER and PGQM are trained end-to-end under this objective without additional supervision. 

\paragraph{Implementation details.}
We adopt RF-DETR with a frozen DINOv3 encoder as the baseline. GAER uses $K=6$ regression experts and a learnable routing prototype bank, while PGQM employs a fixed objectness prototype bank with $M=4$ prototypes. We set $\tau_{\mathrm{box}}=\tau_{\mathrm{obj}}=0.07$ in Eqs.~(\ref{eq:routing_weight}) and~(\ref{eq:cls_proto_prob}), and $\eta=1.1$ in Eq.~(\ref{eq:routed_residual}).
All experiments use distributed data-parallel training on two NVIDIA RTX 3090 GPUs. We train for 48 epochs with a batch size of 4 per GPU and gradient accumulation over 4 steps. AdamW is used with an initial learning rate and weight decay of $1\times10^{-4}$, together with gradient clipping at a maximum norm of 0.1.

\begin{table}[t!]
\centering
\scriptsize
\setlength{\tabcolsep}{1.1pt}
\caption{Quantitative results on the BTCO-Real (Test) (\%).} \vspace{-10pt}
\begin{tabular}{l|ccc|ccc|c|c|c}
\toprule
Methods & AP$_s$ & AP$_m$ & AP$_l$ & AP & AP$_{50}$ & AP$_{75}$  & AR$_{100}$   & Params. (M) &FPS$\uparrow$\\
\midrule
RPN &10.5 &14.5 &7.0 &10.8 &26.2 &7.5  &44.2  &27.5 (27.2)  &49.5 \\
CAOD &17.1 &25.6 &28.7 &23.4 &49.2 &19.7  &38.1   &41.3 (41.1)  &5.78   \\
MAVL &4.4 &18.4 &41.6 &20.8 &31.7 &22.2   &39.3   &59.0 (58.8)  &19.7  \\
DiPEx &16.3 &29.5 &35.6 &26.6 &36.1 &28.6   &39.7  &172.8 (*)  &8.0   \\
PF-RPN &13.5 &32.0 &38.0 &26.4 &34.9 &29.0  &48.5   &135.6 (78.3)  &9.4   \\
RF-DETR $_{\text{(DINOv2-S)}}$ &13.5 &23.5 &26.8 &20.4 &39.6 &18.8   &53.1  &135.6 (135.6)  &50.7   \\\hline
RF-DETR $_{\text{(DINOv3-L)}}$ &14.2 &23.1 &22.3 &24.5 &44.5 &20.8   &53.0  &315.1 (11.9)  &18.7   \\\hline
\textbf{ECADet} $_{\text{(DINOv3-S)}}$ &13.4 &33.1 &47.5 &30.1 &54.4 &29.1  &52.3  &32.2 (10.6)  &44.6   \\
\textbf{ECADet} $_{\text{(DINOv3-B)}}$ &19.6 &39.2 &51.7 &36.0 &60.9 &36.1  &58.0  &98.2 (12.6)  &42.3   \\
\textbf{ECADet} $_{\text{(DINOv3-L)}}$ &25.5  &44.7  &55.0  &41.0 &65.4  &42.7  &61.9   &316.1 (12.9)  &17.3   \\
\bottomrule
\end{tabular}
\vspace{-2pt}
\begin{flushleft}
\scriptsize
\textit{Note:} 
The `*' denotes 256 trainable prompt parameters.
\end{flushleft}
\vspace{-17pt}
\label{tab:BTCO-Real}
\end{table}

\section{Experimental}
\subsection{Setup}

\textbf{Settings.}
All methods are evaluated under the ECAD setting, ignoring category labels and treating all annotated visual elements as foreground. They predict category-agnostic boxes ranked by objectness confidence without class-specific post-processing. 
ECADet builds on RF-DETR with a frozen DINOv3 encoder and is trained on BTCO-Real (Train) and evaluated on BTCO-Real (Test) and BTCO-XDomain.

\paragraph{Comparison methods.}
We compare ECADet with RPN~\cite{ren2016faster}, CAOD~\cite{jaiswal2021class}, MAVL~\cite{maaz2022class}, DiPEx~\cite{syuen2024dipex}, PF-RPN~\cite{tang2026prompt}, and two RF-DETR baselines using a trainable DINOv2 (original RF-DETR) or frozen DINOv3 encoder. These methods cover region proposal, CAD, vision-language-assisted discovery, and prompt-free proposal generation. RPN, CAOD, DiPEx, both RF-DETR baselines, and ECADet are trained on BTCO-Real (Train), while MAVL and PF-RPN use publicly released models trained on larger-scale datasets. For methods with class-specific predictions, category labels are discarded and confidence scores are treated as objectness scores under the unified class-agnostic protocol.

\paragraph{Evaluation protocol and metrics.}
All methods are evaluated as single-foreground-class detectors. We report COCO-style AP averaged over IoU thresholds from 0.50 to 0.95 as the primary metric, together with AP$_{50}$, AP$_{75}$, AP$_s$, AP$_m$, and AP$_l$. We also report total parameters (trainable parameters in parentheses) and FPS, measured on a single NVIDIA RTX 3090 GPU with a batch size of 1, excluding data loading and post-processing.
For proposal-generation methods, we additionally report AR$_K$ using the top-$K$ predictions ranked by objectness. This fixed budget prevents inflated recall from excessive proposals and ensures fair comparison. Unless otherwise specified, we set $K=100$ and report AR$_{100}$.

\subsection{Main Results}

\paragraph{Results on BTCO-Real (Test).}
As shown in Table~\ref{tab:BTCO-Real}, ECADet consistently outperforms all comparison methods. With DINOv3-L, it achieves 41.0 AP, exceeding the strongest competitor, DiPEx, by 14.4 points. It also attains the best AP$_{50}$, AP$_{75}$, and AR$_{100}$, surpassing the corresponding best competing results by 16.2, 13.7, and 8.8 points, respectively. Consistent gains across object scales demonstrate its ability to capture diverse expanded objectness.
ECADet also offers favorable accuracy--efficiency trade-offs across backbone sizes. The DINOv3-S variant reaches 30.1 AP at 44.6 FPS with only 32.2M parameters, outperforming all competitors. Compared with the RF-DETR (DINOv3-L) baseline, ECADet improves AP from 24.5 to 41.0 while adding only 1.0M trainable parameters and incurring modest inference overhead, validating the effectiveness of GAER and PGQM.

\paragraph{Results on BTCO-XDomain.}
As shown in Table~\ref{tab:BTCO-XDomain}, all ECADet variants achieve the best AP, demonstrating strong cross-domain robustness and favorable scaling with backbone size. Even with DINOv3-S, ECADet reaches 45.8 AP, outperforming PF-RPN by 2.9 points; DINOv3-L further raises AP to 65.1, yielding a 22.2-point gain. It also achieves the best AP$_{50}$, AP$_{75}$, and AR$_{100}$, exceeding the strongest competing results by 21.5, 19.3, and 9.3 points, respectively. Consistent gains across all scales indicate effective expanded objectness modeling under appearance shifts. These results show that GAER and PGQM generalize well to sketches, paintings, cartoons, and adverse-condition scenes rather than overfitting domain-specific appearance cues.

\begin{table}[t!]
\centering
\scriptsize
\setlength{\tabcolsep}{1.5pt}
\caption{Quantitative results on the BTCO-XDomain (\%).} \vspace{-10pt}  
\begin{tabular}{l|ccc|ccc|c}
\toprule
Methods & AP$_s$ & AP$_m$ & AP$_l$ & AP & AP$_{50}$ & AP$_{75}$  & AR$_{100}$ \\
\midrule
RPN &5.5 &13.6 &7.3 &8.8 &20.4 &6.5  &57.5    \\
CAOD &11.7 &25.7 &30.0 &27.4 &54.0 &25.0   &50.7    \\
MAVL &5.9 &20.7 &46.3 &36.1 &50.2 &39.3   &69.6    \\
DiPEx &2.9 &9.0 &29.8 &22.2 &33.6 &22.0    &40.3   \\
PF-RPN &17.1 &36.7 &49.8 &42.9 &54.1 &47.0    &82.4   \\
RF-DETR $_{\text{(DINOv2-S)}}$ &14.8 &24.9 &25.2 &23.9 &37.1 &25.3   &77.0   \\\hline
RF-DETR $_{\text{(DINOv3-L)}}$ &24.1 &30.7 &24.9 &26.0 &37.6 &28.6   &80.4   \\\hline
\textbf{ECADet} $_{\text{(DINOv3-S)}}$ &21.5 &42.4 &50.6 &45.8 &65.3 &49.3   &79.9    \\
\textbf{ECADet} $_{\text{(DINOv3-B)}}$ &28.8 &49.4 &53.5 &50.0 &68.7 &53.8   &83.4    \\
\textbf{ECADet} $_{\text{(DINOv3-L)}}$ &57.5 &71.4 &64.6 &65.1 &75.6 &66.3   &91.7    \\
\bottomrule
\end{tabular}\vspace{-17pt}
\label{tab:BTCO-XDomain}
\end{table}

\subsection{Component Ablation}
Table~\ref{tab:abla} reports the component ablation of GAER and PGQM on BTCO-Real (Test). Both modules improve the RF-DETR baseline but exhibit distinct effects. GAER raises AP from 24.5 to 27.7, with larger gains in AP$_{50}$ and AR$_{100}$, indicating that geometry-aware expert routing mainly improves localization quality and proposal coverage across diverse visual elements. PGQM achieves a slightly higher AP of 28.0, showing that prototype-guided query modulation enhances objectness representation and ranking reliability. Combining both modules yields 41.0 AP, a 16.5-point gain over the baseline and a substantial improvement over either module alone. It also reaches 65.4 AP$_{50}$ and 42.7 AP$_{75}$, with pronounced gains on medium and large visual elements. These results confirm the complementary roles and strong synergy of GAER and PGQM.

\begin{table}[t!]
\centering
\scriptsize
\setlength{\tabcolsep}{1.2pt}
\caption{Ablation study of GAER and PGQM (\%).} \vspace{-10pt}  
\begin{tabular}{l|ccc|ccc|c}
\toprule
Method & AP$_s$ & AP$_m$ & AP$_l$ & AP & AP$_{50}$ & AP$_{75}$  & AR$_{100}$ \\
\midrule
RF-DETR $_{\text{(DINOv3-L)}}$ &14.2 &23.1 &22.3 &24.5 &44.5 &20.8  &53.0    \\\hline
~+ GAER &19.3 &30.5 &31.7 &27.7 &51.2 &26.5  &59.5    \\
~+ PGQM &18.8 &29.7 &31.7 &28.0 &49.0 &25.8  &58.5    \\
~+ GAER + PGQM (ECADet $_{\text{(DINOv3-L)}}$) &25.5 &44.7 &55.0 &\textbf{41.0} &65.4 &42.7 &61.9  \\
\bottomrule
\end{tabular}\vspace{-7pt}
\label{tab:abla}
\end{table}

\subsection{Structural Hyperparameter Analysis}
We analyze the structural configurations of GAER and PGQM under single-module settings. The number of regression experts is studied with RF-DETR+GAER at $\eta=0.5$, while the number of offline objectness prototypes is examined with RF-DETR+PGQM. Based on the results, we set the numbers of experts and prototypes to 6 and 4, respectively, and evaluate the sensitivity to $\eta$ using the full ECADet.  

\paragraph{Effect of the number of regression experts.}
Table~\ref{tab:exp_num} investigates the effect of the number of regression experts in GAER on BTCO-Real (Test). Overall AP first improves and then declines as the expert count increases. Using 6 experts achieves the best AP of 27.7, outperforming the other configurations. This suggests that a moderate expert count provides sufficient capacity to model diverse geometric patterns while maintaining effective specialization. Too many experts may reduce the training samples assigned to each expert, leading to under-trained experts and less reliable routing. Therefore, we adopt 6 regression experts.

\begin{table}[t!]
\centering
\scriptsize
\setlength{\tabcolsep}{1.5pt}
\caption{Effect of regression expert count in GAER (\%).}\vspace{-10pt} 
\begin{tabular}{l|ccc|ccc|c}
\toprule
\#Reg. Exp. & AP$_s$ & AP$_m$ & AP$_l$ & AP & AP$_{50}$ & AP$_{75}$  & AR$_{100}$ \\
\midrule
5 &19.5 &29.7 &30.6 &26.0 &45.3 &25.9  &59.5    \\
\underline{6} &19.3 &30.5 &31.7 &\textbf{27.7} &51.2 &26.5  &59.5    \\
7 &18.3 &28.1 &29.3 &24.5 &43.7 &24.0  &58.6    \\
8 &20.1 &30.9 &30.5 &23.5 &43.3 &25.3  &59.9    \\
9 &17.7 &27.1 &27.8 &23.5 &43.2 &22.7  &57.3    \\
\bottomrule
\end{tabular}\vspace{-7pt}
\label{tab:exp_num}
\end{table}

\begin{table}[t!]
\centering
\scriptsize
\setlength{\tabcolsep}{1.5pt}
\caption{Effect of objectness prototype count in PGQM (\%).} \vspace{-10pt}
\begin{tabular}{l|ccc|ccc|c}
\toprule
\#Obj. Proto. & AP$_s$ & AP$_m$ & AP$_l$ & AP & AP$_{50}$ & AP$_{75}$  & AR$_{100}$ \\
\midrule
3 &20.1 &31.5 &33.1 &27.6 &48.0 &27.6  &59.4    \\
\underline{4} &18.8 &29.7 &31.7 &\textbf{28.0} &49.0 &25.8 &58.5 \\
5 &18.7 &28.4 &30.7 &25.2 &44.9 &24.8  &58.3    \\
6 &17.9 &27.8 &28.9 &24.1 &43.4 &23.6  &58.0    \\
\bottomrule
\end{tabular}\vspace{-7pt}
\label{tab:proto_num}
\end{table}

\begin{table}[t!]
\centering
\scriptsize
\setlength{\tabcolsep}{1.5pt}
\caption{Sensitivity to the routed residual scale $\eta$ (\%).} \vspace{-10pt}  
\begin{tabular}{l|ccc|ccc|c}
\toprule
$\eta$ & AP$_s$ & AP$_m$ & AP$_l$ & AP & AP$_{50}$ & AP$_{75}$  & AR$_{100}$ \\
\midrule
0.1 &22.7&41.4&47.4&36.0&62.0&36.7&59.2   \\
0.3 &23.8&42.0&51.4&38.4&63.0&39.8&59.2   \\
0.5 &25.5&44.0&52.2&39.7&64.2&41.2&62.0   \\
0.7 &25.4&43.8&53.4&40.0&64.6&41.4&62.1   \\
0.9 &25.5&44.6&54.7&40.8&65.3&42.4&61.8   \\
\underline{1.1} &25.5&44.7&55.0&\textbf{41.0}&65.4&42.7&61.9 \\
1.3 &24.4&41.5&49.0&37.4&61.0&38.9&61.6   \\
\bottomrule
\end{tabular}\vspace{-15pt}
\label{tab:eta}
\end{table}

\paragraph{Effect of the number of offline objectness prototypes.}
Table~\ref{tab:proto_num} investigates the number of offline objectness prototypes in PGQM on BTCO-Real (Test). Using 4 prototypes achieves the best AP of 28.0 and the highest AP$_{50}$ of 49.0. Although 3 prototypes yield slightly higher scale-specific APs, AP$_{75}$, and AR$_{100}$, the 4-prototype setting offers the best overall accuracy, suggesting a better balance between prototype diversity and matching stability. Increasing the count beyond 4 consistently degrades performance, likely because excessive partitioning produces redundant or weakly supported prototypes, reducing matching reliability. Therefore, we adopt 4 offline objectness prototypes in PGQM.

\paragraph{Sensitivity to the routed residual scale $\eta$.}
Table~\ref{tab:eta} analyzes the sensitivity to $\eta$ using the full ECADet with 6 regression experts and 4 offline objectness prototypes. As $\eta$ increases from 0.1 to 1.1, AP generally improves from 36.0 to 41.0, indicating that an appropriate residual scale allows the expert-routed branch to complement the base box prediction. The best performance is achieved at $\eta=1.1$, with 41.0 AP, 65.4 AP$_{50}$, and 42.7 AP$_{75}$. Further increasing $\eta$ to 1.3 causes a drop, suggesting that excessive residual refinement may perturb the base prediction and reduce localization stability. Meanwhile, the stable performance within $\eta\in[0.7,1.1]$ shows limited sensitivity to moderate variations. Therefore, we set $\eta=1.1$ in the final model.

\subsection{Qualitative Results}

\begin{figure}[t!]
	\centering
	\includegraphics[width=1.0\linewidth]{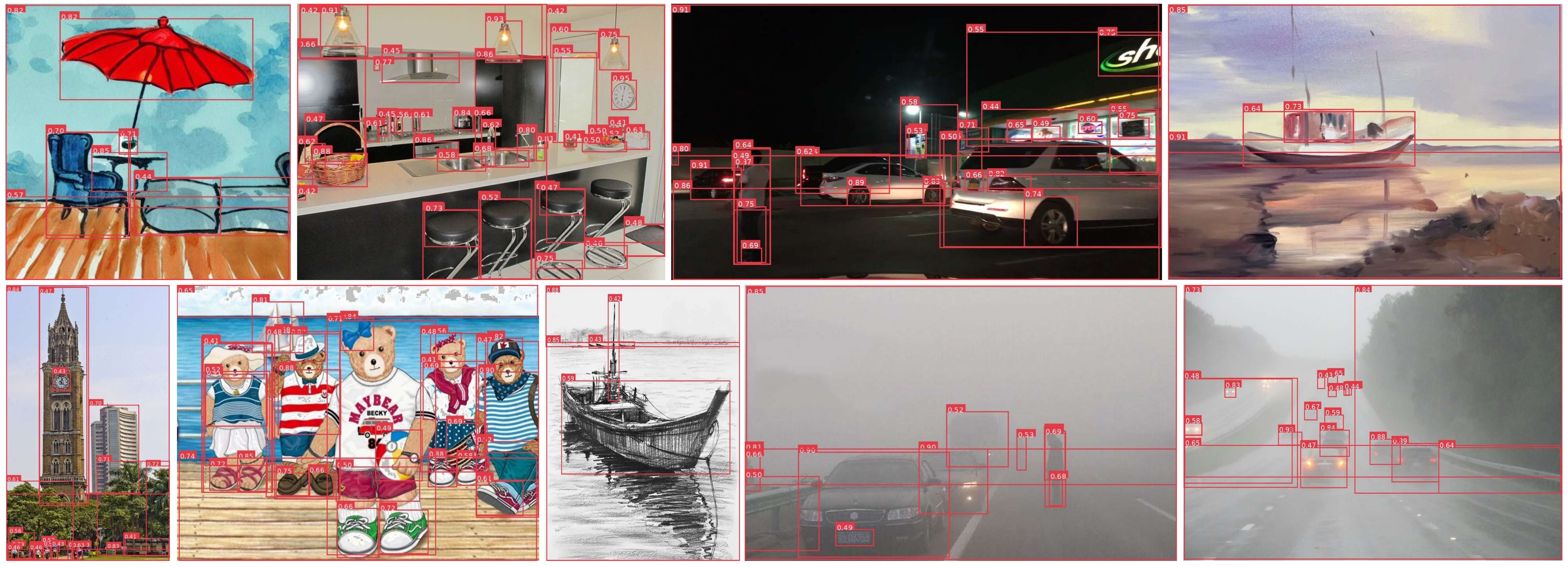}
	\vspace{-19pt}
    \caption{Qualitative results of ECADet on BTCO-Bench.} 
	\label{fig:results_main}
	\vspace{-17pt}
\end{figure}

Fig.~\ref{fig:results_main} presents representative ECADet predictions on BTCO-Real and BTCO-XDomain, covering real, pencil sketches, cartoons, oil paintings, watercolors, and adverse-condition scenes. Across diverse domains, ECADet produces complete and well-aligned boxes for both conventional thing instances and semantically meaningful stuff-like regions, such as sky, water, floors, and roads, while maintaining low prediction redundancy under substantial appearance shifts. Further qualitative comparisons are in the \textit{\textbf{supplementary material}}.

\section{Conclusion}

In this work, we introduce Expanded Class-Agnostic Detection (ECAD), extending thing-centric objectness to semantically meaningful visual elements. We construct BTCO-Bench for real-world and cross-domain evaluation and propose ECADet, a lightweight detector built on a frozen VLM. ECADet employs GAER and PGQM to model geometric diversity and improve objectness estimation. Extensive experiments demonstrate consistent gains over existing class-agnostic detection and proposal-generation methods, together with strong cross-domain generalization and favorable efficiency. We hope this work inspires broader exploration of objectness beyond conventional instances and advances visual perception in open and diverse environments.


\bibliography{aaai2027}


\end{document}